# Autonomous intelligence for mining productivity: a multi-agent AI approach to operational value creation

N Guerin[1], A Dehoux[2] and M Nakhla[3]

1. Data Scientist – PhD Candidate, Eramet, Paris, 75015, France. Email: nicolas.guerin@minesparis.psl.eu
2. Lead Data Scientist, Eramet, Paris, 75015, France. Email: anita.dehoux@eramet.com
3. Professor, Mines Paris PSL – CGS, Paris, 75006, France. Email: michel.nakhla@minesparis.psl.eu

## ABSTRACT

Integrated Remote Operation Centres (IROCs) are emerging as the architectural frontier of Industry 4.0 for capital-intensive sectors such as mining, oil and gas, and energy. Yet two intertwined gaps constrain their evolution towards full autonomy. First, no integrated agentic architecture has been proposed for mining IROCs that orchestrates Supervisory Control and Data Acquisition systems, digital twins, predictive analytics and human–machine interfaces under safety-critical constraints. Second, no evaluation framework jointly addresses the real-time, Health–Safety–Environment, regulatory-auditability and cost-per-decision requirements that govern industrial agentic deployments. This paper makes four contributions. It synthesises industry needs and quantified gains documented in analogous industries that motivate an agentic approach to mining IROCs. It introduces EI2ROC, a three-tier multi-agent architecture combining a Large Language Model orchestrator with specialised monitoring and production-capacity-prediction agents communicating through a Model Context Protocol–style tool interface. It provides a critical review of agentic evaluation along six orthogonal dimensions and identifies the industrial blind spot of current benchmarks. Finally, it proposes a multi-dimensional benchmarking framework for industrial agentic AI comprising scenario library, harness, metrics and human-calibration protocol. The paper states explicitly which components are implemented in the current prototype, which belong to the proposed reference design and which remain future work, and reports preliminary cost and latency observations from an operating site. Together, EI2ROC and the benchmarking framework constitute a methodological blueprint transferable across CAPEX-intensive industries.

## INTRODUCTION

### 1.1 Macro context

Capital-intensive industries (mining, oil and gas, and energy) face mounting pressure to optimise asset utilisation, reduce operational expenditure and maintain safety standards in increasingly remote and hazardous environments. Digitalisation has emerged as a strategic imperative in these sectors, driven by the need to sustain competitiveness amid volatile commodity markets, ageing infrastructure and stringent regulatory requirements (Hermann, Pentek and Otto, 2016; Litvinenko et al, 2022). In mining, where new large-scale operations often require billions of dollars in capital investment, the economic case for automation and remote monitoring is particularly acute (Litvinenko et al, 2022). Digital transformation initiatives in analogous industries have documented operational efficiency gains: in oil and gas, integrated monitoring systems combining digital twins and cloud computing have, in analogous industries, been associated with reductions in maintenance time and infrastructure costs (Shekhawat, Barua and Bhatia, 2025), and Mining 4.0 technologies (encompassing artificial intelligence, big-data analytics and smart sensors) have demonstrated productivity gains through optimised machine work and enhanced decision-making (Zhironkin and Ezdina, 2023).

## 1.2 The open-pit mining case

Central to this transformation is the concept of Integrated Remote Operation Centres (IROCs): centralised control hubs that enable real-time monitoring, coordination and decision-making across geographically dispersed mining assets. IROCs consolidate data streams from sensors, equipment and personnel into unified platforms that support predictive maintenance, scenario simulation and strategic planning (Chen et al, 2021). Yet structural uncertainties (weather variability, mechanical breakdowns, ore-body heterogeneity) compound dispatcher cognitive load and create decision latency between event detection and corrective action; recent work on probabilistic hauling-fleet capacity prediction quantifies these uncertainties and exemplifies the kind of analytics that an agentic IROC could orchestrate (Guerin et al, 2025).

## 1.3 The promise of agentic AI for IROCs

Multi-Agent Systems (MAS) and the more recent generation of Large Language Model (LLM)-orchestrated agentic AI offer a credible path beyond current IROC capabilities. Classical MAS deployments in manufacturing (Cruz Salazar and Vogel-Heuser, 2022; Seitz et al, 2021) and hybrid agentic–MAS architectures in smart manufacturing (Farahani, Khan and Wuest, 2026) and offshore oil-and-gas safety-critical operations (Shekhawat, Barua and Bhatia, 2025) demonstrate the maturity of the underlying technologies, while ReAct-style and Reflexion-style agents (Yao et al, 2023; Shinn et al, 2023) show that LLMs can be coupled with deterministic tools to deliver verifiable, auditable behaviour. The transposition of these patterns to mining IROCs remains, however, an open research question.

## 1.4 Research gaps and contributions

Two intertwined scientific gaps constrain the realisation of fully autonomous mining IROCs. Gap (i), Architecture. No integrated agentic architecture has been proposed for mining IROCs that orchestrates SCADA, digital twins, predictive analytics and human–machine interfaces under safety-critical constraints; existing mining-side AI contributions remain isolated analytics services rather than coordinated agent societies (Guerin et al, 2025). Gap (ii), Evaluation. While balanced agent-evaluation frameworks have emerged (Michelakis, Hadjiyiannis and Stamoulis, 2025; Kapoor et al, 2024; Lee et al, 2025; Theodorakopoulos and Theodoropoulou, 2026; Radanliev, 2026), none jointly accommodates the real-time, HSE, regulatory-auditability and cost-per-decision constraints characteristic of safety-critical industrial agents.

To address these gaps, this paper makes four contributions: (i) a synthesis of industry needs and of the gains documented in analogous industries that motivate an agentic approach to mining IROCs (Section 3); (ii) EI2ROC, a three-tier LLM-orchestrated multi-agent architecture answering Gap (i), detailed down to its agent routing, tool-call protocol and end-to-end query lifecycle (Section 4); (iii) a critical review of agentic-AI benchmarking along six orthogonal dimensions, identifying the industrial blind spot of current frameworks; and (iv) a multi-dimensional benchmarking framework for industrial agentic AI (scenario library, harness, metrics and calibration protocol) answering Gap (ii) (Section 5). Throughout, implemented components are distinguished from proposed reference design and from future work (Section 4.10), and the preliminary observations of Section 4.11 characterise the operating envelope of the prototype rather than evaluating it; controlled evaluation is the object of the calibration campaign of Section 7.

# BACKGROUND AND RELATED WORK

This section reviews four bodies of literature that frame the contributions of this paper (the architectural evolution from SCADA to IROCs, classical MAS for Industry 4.0, the emerging paradigm of LLM-orchestrated agentic AI, and the nascent landscape of agent-evaluation frameworks) and closes with the two research gaps addressed by the remainder of the paper.

## 2.1 From SCADA to integrated remote operation centres

Industrial control in capital-intensive sectors has progressed through three generations. SCADA and Distributed Control Systems automated individual processes and enabled remote monitoring but operated as isolated subsystems with limited cross-domain interoperability (Chen et al, 2021); centralised control rooms then consolidated monitoring into unified spaces, yet production, equipment-health, environmental and logistics data continued to reside on separate platforms, perpetuating fragmented, schedule-driven decision-making (Chen et al, 2021). IROCs constitute the current frontier: centralised digital platforms that aggregate real-time streams from sensors, mobile equipment and field personnel, expose them through unified analytics, and support remote dispatching, predictive maintenance and scenario simulation across the value chain (Chen et al, 2021; Zhironkin and Ezdina, 2023), relocating supervisory staff to remote hubs without relaxing HSE standards (Chen et al, 2021). Three limitations recur. Data silos and weak semantic interoperability persist even within nominally integrated platforms, mirroring offshore oil-and-gas operations (Shekhawat, Barua and Bhatia, 2025). Decision-making remains predominantly reactive in the absence of predictive analytics able to anticipate degradation (Chen et al, 2021). And IROC workflows exhibit heavy operator dependency, as dispatchers synthesise multiple dashboards under time pressure (Guerin et al, 2025). Cyber-security exposure, lack of standardisation and skills shortages are repeatedly identified as further deployment barriers (Shekhawat, Barua and Bhatia, 2025). The quantitative gains that motivate addressing these limitations are documented in analogous industries and reviewed in Section 3.2.

## 2.2 Multi-agent systems for Industry 4.0

Classical MAS emerged in the mid-1980s as a distributed paradigm in which autonomous agents coordinate through explicit protocols and shared ontologies (Wooldridge, 2009), with the Belief–Desire–Intention model as the canonical cognitive architecture (Rao and Georgeff, 1995; Bordini, Hübner and Wooldridge, 2007) and FIPA-ACL standardising interoperability (Poslad, 2007). Industrial deployments have delivered tangible benefits: holonic control distributes intelligence across cell-, line- and plant-level agents, enabling negotiation-based task allocation (Cruz Salazar and Vogel-Heuser, 2022); and platform-independent agent architectures coupled with OPC UA deliver hardware-agnostic integration and scale to robust networks of production resources (Seitz et al, 2021). Yet classical MAS have not been deployed end-to-end in mining IROCs: capacity prediction, dispatching and maintenance remain separate analytics services rather than an orchestrated agent society (Guerin et al, 2025), leaving their modularity, reconfigurability and distributed intelligence largely untapped on the mining side.

## 2.3 The rise of LLM-orchestrated agentic AI

LLM-powered agents mark a shift from static, request-driven functions towards dynamic, goal-oriented multi-agent ecosystems (Cao and Yu, 2026). Two patterns are now foundational: ReAct interleaves reasoning traces with action execution (Yao et al, 2023), and Reflexion augments agents through linguistic self-feedback stored in episodic memory (Shinn et al, 2023); building on these primitives, the orchestrator/sub-agent pattern decomposes high-level tasks into specialised roles coordinated by a central planner (Farahani, Khan and Wuest, 2026). Hybrid agentic–MAS architectures are reaching industry: prescriptive-maintenance frameworks combine an LLM planner with rule-based and small-language-model agents at the edge (Farahani, Khan and Wuest, 2026), and offshore oil-and-gas deployments integrate agentic decision support into HSE-regulated operations (Shekhawat, Barua and Bhatia, 2025). The Model Context Protocol (MCP) standardises how agents access tools and contextual data through explicit capability contracts (Anthropic, 2024), though forensic analysis of deployed MCP systems shows that reconstructing which tool was called, with which arguments and against which server is often impossible from production logs alone, which places explicit demands on architectural design (Satter et al, 2026). This transition is an architectural rupture (learned probabilistic reasoning replaces rule-based determinism, emergent natural-language collaboration replaces explicit FIPA-ACL coordination, and the locus of intelligence migrates from platform to model) that directly affects verifiability, certification and operator trust. To the authors' knowledge, no published work transposes the orchestrator/sub-agent pattern to mining IROCs, despite an operational profile (heterogeneous subsystems, real-time constraints, regulated decisions) analogous to the manufacturing and offshore cases above.

### 2.4 Evaluating agentic systems: an emerging concern

Classical machine-learning metrics (accuracy, precision, recall, F1) assume a static input–output mapping and fail to characterise agents operating through extended trajectories of perception, planning, tool use and self-correction (Cao and Yu, 2026). Three mismatches are now broadly acknowledged: path correctness, not endpoint success, must be assessed, because agents with identical completion rates may differ in safety and resource use (Michelakis, Hadjiyiannis and Stamoulis, 2025); the non-deterministic nature of LLM reasoning yields multiple valid trajectories, ruling out reference-replication metrics; and process quality, including constraint adherence and explainability, cannot be inferred from outcome correctness alone (Kapoor et al, 2024). A first generation of balanced frameworks has responded, yet none jointly addresses the four constraints that govern CAPEX-intensive deployment: latency reported against declared bounds, HSE compliance and fail-safe behaviour, regulatory auditability against standards such as IEC 61508 and ISO 31000, and cost-per-decision. A detailed review of these frameworks and the resulting industrial blind spot is deferred to Section 5.

### 2.5 Research gap

Two intertwined gaps emerge from this review, formalised in Section 1.4. Gap (i), Architecture. Classical MAS (Cruz Salazar and Vogel-Heuser, 2022; Seitz et al, 2021) and hybrid LLM/agentic-MAS architectures (Farahani, Khan and Wuest, 2026; Shekhawat, Barua and Bhatia, 2025) have been demonstrated in manufacturing and offshore, yet no integrated agentic architecture orchestrates SCADA, digital twins, predictive analytics and human–machine interfaces for mining IROCs under safety-critical constraints; existing mining-side contributions remain isolated analytics services rather than coordinated agent societies (Guerin et al, 2025). Gap (ii), Evaluation. No existing framework jointly accommodates the real-time, HSE, auditability and cost-per-decision constraints of safety-critical industrial agents (Michelakis, Hadjiyiannis and Stamoulis, 2025; Kapoor et al, 2024; Theodorakopoulos and Theodoropoulou, 2026; Radanliev, 2026). Section 3 formalises the operational pain points and the documented stakes; Section 4 presents the EI2ROC architecture answering Gap (i); and Section 5 develops the multi-dimensional benchmarking framework answering Gap (ii).

## INDUSTRY NEEDS AND DOCUMENTED GAINS

### 3.1 Operational pain points in open-pit mining

Open-pit operations face three families of operational pain points that no current IROC fully addresses. Geological variability and grade uncertainty propagate downstream into haulage and processing decisions and force planners to commit equipment under partially observable ore-body knowledge (Guerin et al, 2025). Weather-driven productivity losses are routinely cited as a major source of unplanned downtime in surface mining, and high-resolution rainfall measurements have been shown to be among the most predictive features for short-term fleet capacity (Guerin et al, 2025). Dispatcher cognitive load is a third pain point: operators synthesise information from multiple dashboards (Fleet Management Systems, SCADA, IoT sensors, weather feeds) and make time-critical decisions under uncertainty, a cognitive burden that scales with operational complexity and is amplified by the absence of decision-support that anticipates degradation or process anomalies (Guerin et al, 2025). The latency between event detection and corrective action is itself a performance variable, minutes-scale rather than hours-scale, and is not measured by today's IROC dashboards.

### 3.2 Documented AI / MAS value in analogous industries

Quantitative gains from agentic and MAS approaches are documented in analogous industries. Table 1 summarises representative case studies, drawn from offshore oil and gas (Shekhawat, Barua and Bhatia, 2025), mining predictive maintenance (Chambi et al, 2025), anomaly detection on mining haul-truck engines (Lima and Cutipa Arapa, 2024), hybrid agentic–MAS frameworks for smart manufacturing (Farahani, Khan and Wuest, 2026) and probabilistic capacity prediction on the mining side (Guerin et al, 2025). All quantitative gains reported are documented in analogous industries and prior work and are reproduced here as evidence of the value of agentic / MAS approaches, never as results of the present study.

| Industry | Use case | Documented gain (in analogous industries) | Reference |
|---|---|---|---|
| Offshore oil and gas | Multi-agent orchestration for monitoring, root-cause analysis and predictive maintenance | Substantial reduction in manual intervention time | (Shekhawat, Barua and Bhatia, 2025) |
| Mining (underground) | AI-based predictive maintenance on mining equipment | 8% reduction in maintenance cost and 10% increase in availability | (Chambi et al, 2025) |
| Discrete manufacturing | Distributed cyber-physical analytics through cooperative agents | Improved robustness and horizontal scalability of analytics pipelines | (Queiroz et al, 2022) |
| Smart manufacturing | Hybrid agentic + classical MAS architectures for prescriptive maintenance | Increased modularity and online adaptation to disturbances | (Farahani, Khan and Wuest, 2026) |
| Mining (production capacity) | Deep-learning model trained on real and simulated fleet data for capacity forecasting | Median absolute error (MedAE) < 15% on short-horizon capacity prediction | (Guerin et al, 2025) |

Table 1: Documented Gains from AI / MAS Deployments in Analogous Industries

### 3.3 Functional requirements for a mining IROC

The above pain points translate into five functional requirements. (i) Hierarchical performance monitoring through a Value Driver Tree (VDT) linking operational Key Performance Indicators (KPIs) to business outcomes. (ii) Overall Equipment Effectiveness (OEE = availability × utilisation × performance × quality) computed continuously and benchmarked across shifts and sites. (iii) Short-Interval Control (SIC) loops, typically of one to two hours, that identify drifts and trigger corrective actions either automatically or under human validation. (iv) Planning under uncertainty supported by probabilistic capacity prediction with explicit confidence intervals (Guerin et al, 2025). (v) A conversational interface that lets dispatchers query the agentic stack in natural language.

### 3.4 Non-functional requirements

Six non-functional requirements govern industrial deployability. Explainability: every recommendation must reference its causal trace. Auditability: the decision provenance must be reconstructible against regulatory standards (IEC, 2010; ISO, 2018). Latency: three regimes are distinguished and used consistently throughout this paper. Hard real-time control, at millisecond scale, remains the responsibility of the underlying SCADA and programmable-logic layer and lies outside the scope of the agentic tier, which makes no hard real-time claim. The agentic tier operates in a soft real-time regime at minutes-scale, aligned with the Short-Interval-Control loop, in which an answer is expected within minutes rather than hours. Latency is reported as a distribution, through a median and a high percentile such as p95 measured against a stated budget, never as a guaranteed bound. Cost per decision: LLM-orchestration cost must remain a small fraction of the value of the decision. Safety: no destructive action without human-in-the-loop validation. Modularity: the architecture must support the progressive addition of maintenance, energy, safety and logistics agents without rework.

### 3.5 Why evaluation becomes a deployment prerequisite

Functional and non-functional requirements jointly imply that, in an industrial context, an agentic system cannot be deployed unless it can be measured. Cost per decision, auditability and HSE compliance are not optional benchmarks: they are deployment gates. Section 5 shows that no existing benchmarking framework jointly addresses these gates (Gap (ii), Section 1.4) and proposes one designed to close the gap.

## THE EI2ROC ARCHITECTURE

### 4.1 Overview: three-tier multi-agent architecture

EI2ROC organises a mining IROC as a three-tier society of cooperating agents, connected by a native tool-use interface functionally equivalent to MCP (Anthropic, 2024) and illustrated in Figure 1.

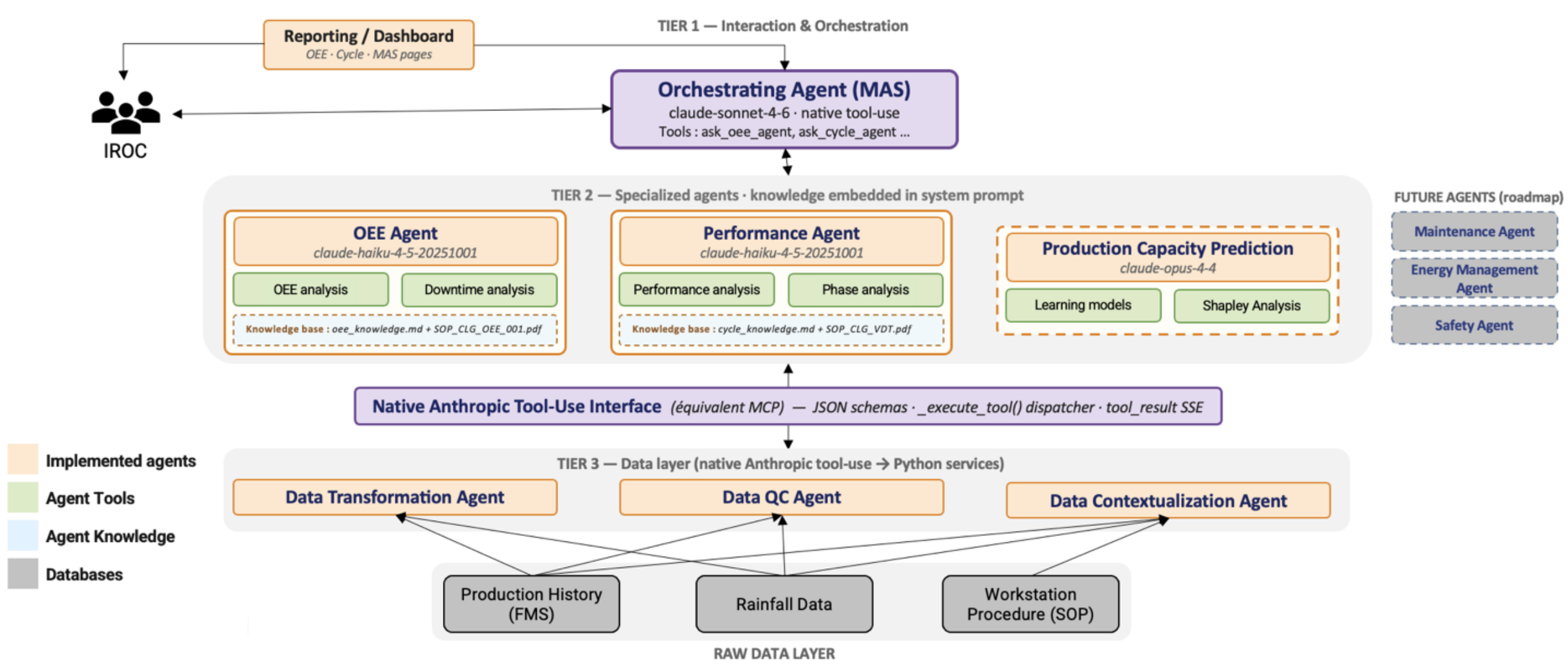

FIG 1: EI2ROC three-tier multi-agent architecture for a mining IROC

Tier 1 (Interaction & Orchestration) hosts a single orchestrating agent that exposes a conversational interface to the IROC and routes work. Tier 2 (Specialised agents) hosts domain experts, namely an OEE agent, a performance (cycle) agent and a production-capacity-prediction agent, each running its own model, its own tool set and its own embedded knowledge, namely a domain knowledge file and the relevant Standard Operating Procedures (SOPs). Tier 3 (Data layer) hosts service agents that transform, quality-control and contextualise the raw sources, namely Fleet-Management-System (FMS) production history, rainfall and other IoT feeds, and workstation SOPs, before any of it reaches the reasoning agents. A reporting and dashboard surface renders results back to operators. Rather than the monolithic analytics service typical of current IROCs, EI2ROC distributes intelligence across specialised, independently versioned agents while preserving a single conversational entry point and an end-to-end audit trail. Model assignment follows a cost–capability rationale: a high-capability reasoning model orchestrates, cost-efficient models serve the high-frequency monitoring agents, and a high-capability model backs prediction (in the reference implementation, a Claude Sonnet-class orchestrator, Claude Haiku-class monitoring agents and a Claude Opus-class capacity agent). Of these three assignments, the orchestrator and the monitoring agents are the ones exercised in the testing environment reported in Section 4.11, where they account for 13 and 20 model calls respectively across the four queries of Table 3.

## 4.2 Tier 1: the orchestrator agent

The orchestrator is the IROC's single point of contact. It receives natural-language queries, decomposes operator intent into sub-tasks, delegates them to Tier-2 agents through typed tool calls, handles errors, back-off and retries, and synthesises a traceable answer. It maintains short-term memory (the current context window) and a long-term vector store of prior interactions that supports shift handovers and a growing case base. Crucially, the orchestrator carries no domain analytics of its own: it owns planning, delegation and synthesis, exposing each child agent as a tool (for example ask_oee_agent or ask_cycle_agent) whose free-text description and JSON schema it reads to decide what to invoke. Every response returned to the dispatcher cites the execution trace of sub-agent invocations that produced it, which is what turns the exchange from a chatbot conversation into an auditable decision record. Figure 2 traces a concrete dispatcher query through this loop, from the natural-language question to the orchestrator's plan, its delegation to the OEE and performance agents, and the trace-cited answer.

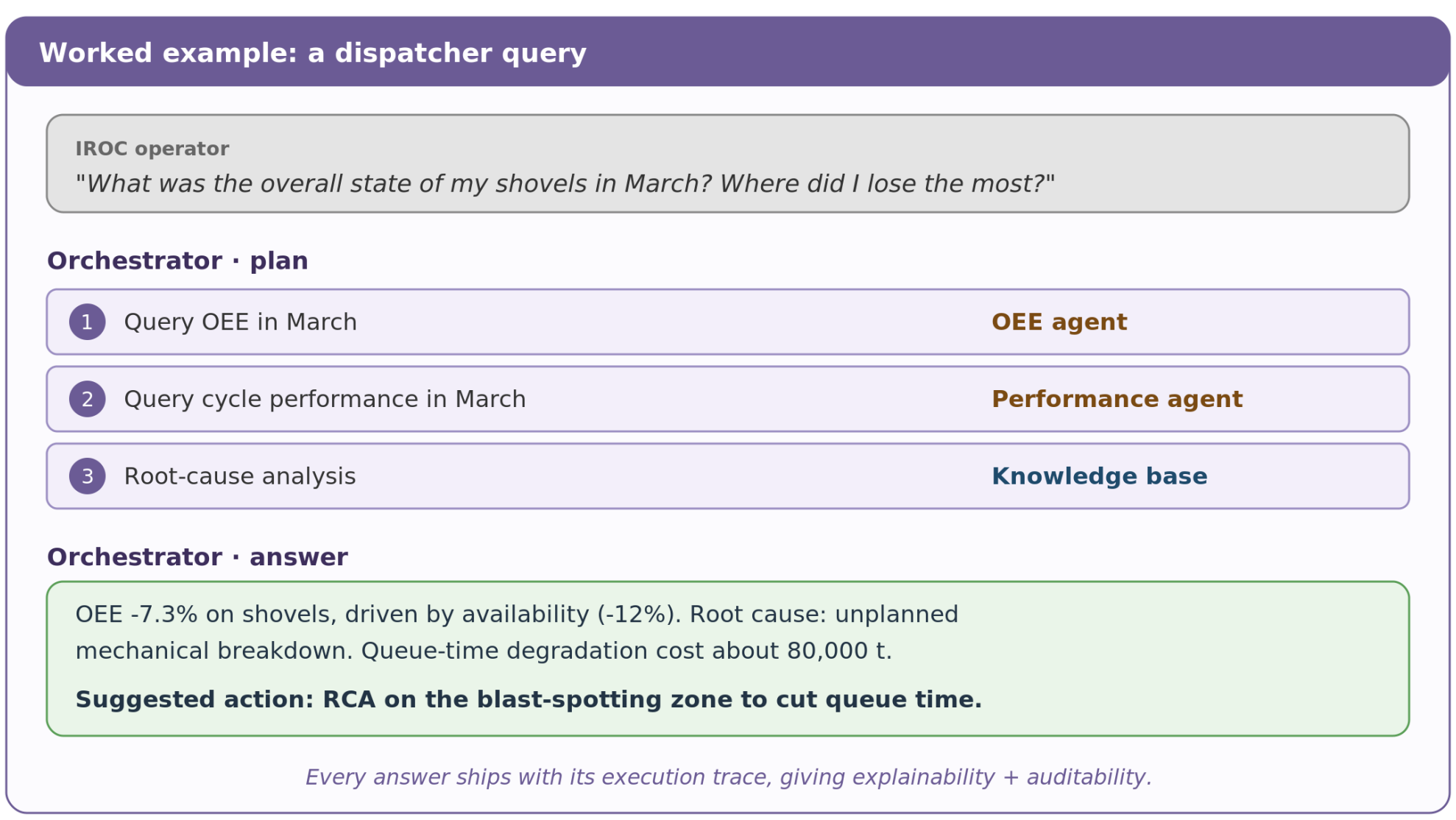

FIG 2: Worked example of a dispatcher query resolved by the orchestrator

## 4.3 Tier 2: specialised domain agents

The second tier comprises three domain agents, each encapsulating its own model, tools, SOPs and knowledge base so that domain expertise lives in the agent rather than in the orchestrator. The OEE agent computes Overall Equipment Effectiveness decomposed by asset and period, and analyses downtime; its tools include compute_oee(), get_downtime_reasons() and query_shift_events(), and its knowledge base pairs an oee_knowledge file with the relevant OEE SOP used by the company. The performance (cycle) agent computes cycle KPIs, namely average cycle duration, productivity (t/h), payload and queue ratio, and runs phase analysis across the cycle's phases to localise bottlenecks; its tools include compute_cycle_metrics(), get_phase_analysis() and query_cycle_data(), backed by a cycle_knowledge file and the VDT SOP, standard of the company. Together these two agents realise the hierarchical Value Driver Tree and the Short-Interval-Control loops of Section 3.3. The production-capacity-prediction agent wraps the deep-learning model of (Guerin et al, 2025), trained on real telemetry and calibrated discrete-event simulation; it exposes predict_next_payload() for probabilistic short-horizon forecasts, an optimise() what-if tool (shovel down, weather event, fleet change, grade shift) and a shap() tool that returns SHAP feature attributions with every forecast, so that predictions are designed to be traceable rather than opaque and to satisfy the explainability requirement of Section 3.4. This agent is integrated into the architecture but is hosted in a separate environment and is not reachable from the orchestrator in the testing configuration reported in Section 4.11. Because each agent is independently versioned, the society can evolve agent by agent without disturbing the others.

### 4.4 Tool-use interface and tool-call protocol

The tiers communicate through a native tool-use interface functionally equivalent to MCP: JSON tool schemas, a server-side dispatcher and streamed tool results (Anthropic, 2024). A single tool call follows three steps (Figure 3). First, the calling model emits a typed tool_use block naming the tool and supplying structured arguments. Second, the dispatcher translates it into a JSON-RPC tools/call and executes the corresponding Python service. Third, the service returns a typed tool_result block, a structured payload carrying metrics, a suspect cause and supporting evidence. Critically, before any result reaches the model it passes a stack of validation layers: schema validation, authentication and scope checks, rate-limiting and append-only audit logging, with response contracts enforced by server-side Pydantic schemas, to which HSE-rule checks are added in the reference design. The reasoning agents therefore only ever see well-typed, policy-checked data. This protocol is where EI2ROC's non-functional requirements (auditability, safety and bounded cost) are designed to be mechanically enforced rather than merely promised; Section 4.10 states which of these layers are active in the current prototype, and establishing that the enforcement holds under adversarial conditions is the object of the framework of Section 5. The protocol incorporates the guardrails recommended by critical evaluations of MCP integration (Satter et al, 2026).

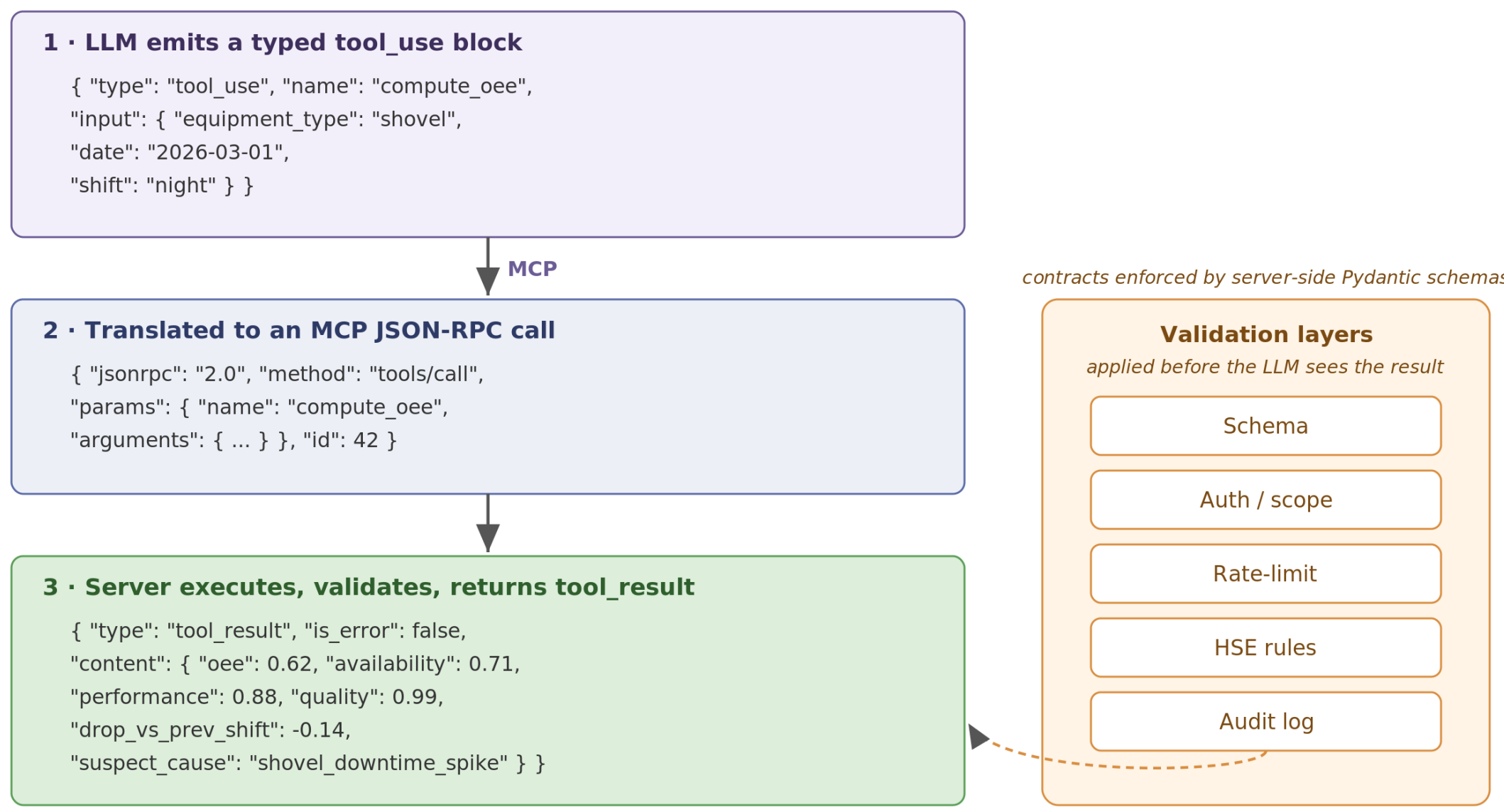

FIG 3: Tool call protocol

## 4.5 Tier 3: the data layer

Beneath the tool interface, a data layer of service agents conditions the raw sources before they are exposed as tools. A data-transformation agent normalises heterogeneous records into common formats; a data-QC agent screens for missing, out-of-range or inconsistent values; and a data-contextualisation agent enriches records with operational context (equipment identity, shift, site, weather window). These agents draw on the raw data layer (FMS production history, rainfall data and workstation SOPs) and address head-on the data-silo and interoperability limitation identified in Section 2.1. By concentrating data conditioning in a dedicated tier, EI2ROC keeps the reasoning agents free of brittle parsing logic and gives the audit trail a single, well-defined point at which data provenance is established.

## 4.6 Agent routing

Routing in EI2ROC is LLM-native: there is no hard-coded switch and no separate intent classifier. The orchestrator reads each tool's free-text description and JSON schema and selects the next action implicitly (Figure 4), mapping a diagnostic question to the monitoring agents, a forward-looking question to the capacity agent, and an actionable request to a reasoning-plus-executor step. It can chain several tool calls within a single turn in a ReAct-style loop (querying the OEE agent, then the cycle agent, then a knowledge step) and applies an early-exit policy, stopping as soon as it has enough evidence to answer or act. Routing is thus adaptive to the query rather than bound to a fixed decision tree, while the typed tool layer keeps every branch observable.

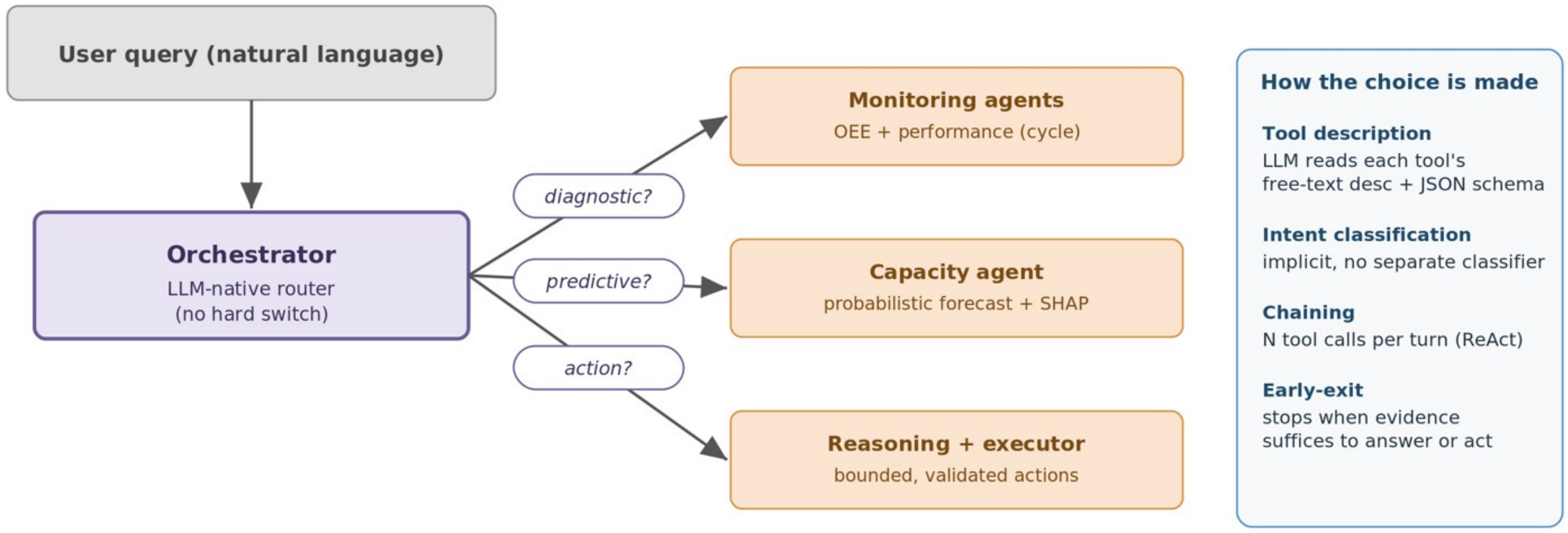

FIG 4: LLM-native agent routing in the orchestrator

## 4.7 End-to-end query lifecycle

A representative diagnostic query illustrates the full loop (Figure 5). (1) The operator asks, in natural language, “Why did performance drop last week?” (2) The orchestrator plans and requests the OEE agent, passing parameters in natural language. (3) The OEE agent issues typed tool calls that fetch FMS/SCADA data, (4) receives the telemetry payload and (5) returns a structured OEE drilldown with a suspect cause. (6) The orchestrator then requests the cycle agent, (7) which fetches cycle data, (8) receives its payload and (9) returns a structured performance report with a suspect cause. (10) The orchestrator synthesises a natural-language answer with a link to the complete execution trace. An illustrative answer, constructed for exposition rather than extracted from a logged run, reads: “OEE −7.3% on shovels, driven by availability (−12%); root cause: unplanned mechanical stoppage; queue-time degradation cost about 80 000 t; suggested action: root-cause analysis on the blast-spotting zone.” This loop is designed to close the detection-to-action gap identified as an unmeasured performance variable in Section 3.1; latency measured on comparable queries in the testing environment is reported in Section 4.11.

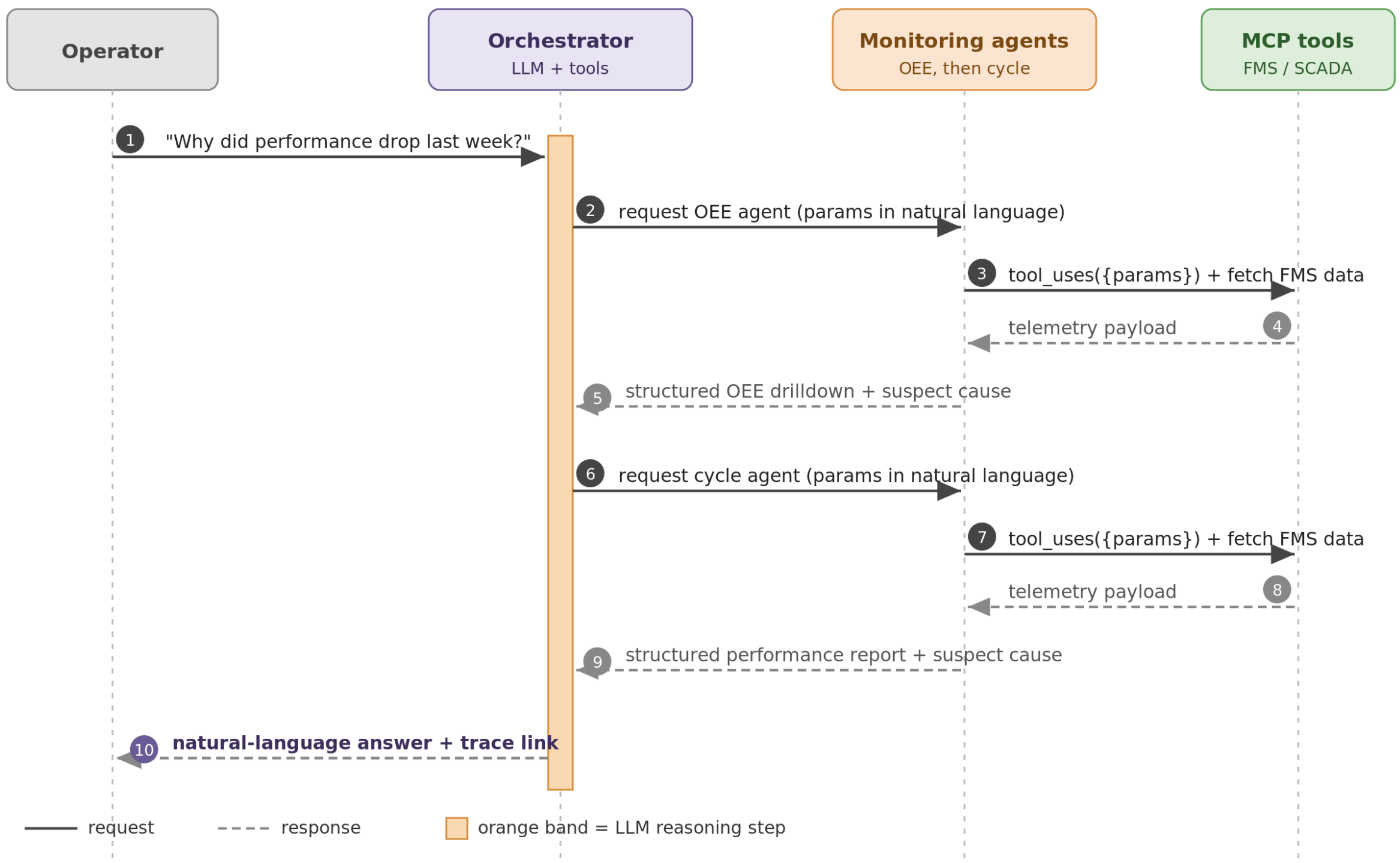

FIG 5: End-to-end query lifecycle across operator, orchestrator, agents and tools

### 4.8 Human–AI collaboration and autonomy levels

EI2ROC exposes three configurable autonomy levels, namely suggestion (the agent proposes, the dispatcher decides), approval (the agent acts on explicit validation) and automatic (the agent acts within a bounded set of pre-authorised, reversible actions), settable per scenario class and per agent. Safety-critical or irreversible actions are escalated to a human dispatcher by construction of the autonomy configuration, which in the current prototype is the mechanism carrying this guarantee, the independent HSE-rule layer of Section 4.4 belonging to the reference design (Section 4.10). Every recommendation references the sub-agent trace that produced it, supporting both regulatory auditability and operator trust (Theodorakopoulos and Theodoropoulou, 2026; Radanliev, 2026).

### 4.9 Deployment, extensibility and scalability

The target deployment couples on-premises infrastructure, for FMS/IoT proximity and low-latency data access, with cloud-side inference for forecasting and orchestration. The agent society is open by construction: roadmap agents for maintenance, energy management and safety can be added as additional Tier-2 experts, each with its own model, tools and knowledge, without refactoring the orchestration backbone or the tool interface. Multi-site federation through federated learning is identified as a longer-horizon direction. This modularity is also what makes EI2ROC measurable agent by agent, the property that the benchmarking framework of Section 5 exploits.

### 4.10 Implementation status

EI2ROC is currently exercised in a testing environment on an operating Eramet site, with a small number of dispatcher users and a single model configuration. Because the components described above sit at different stages of maturity, Tables 2a and 2b state which are implemented and running, which are specified as part of the proposed reference design but not yet active, and which remain future work. The distinction governs how the claims of this paper are to be read: properties attributed to implemented components are observed in operation, whereas properties attributed to reference-design components are design intentions whose verification depends on the framework of Section 5.

Two boundaries deserve emphasis. Within the validation stack of Section 4.4, schema validation, authentication and scope checking, rate limiting and append-only audit logging are active and blocking, whereas the HSE rule checks are specified but not yet enforced at run time, so safety escalation currently rests on the autonomy-level configuration of Section 4.8 rather than on an independent rule layer. Second, the benchmarking harness of Section 5 is not implemented and its scenario library is under construction with site operations experts, which places the calibration campaign of Section 7 as the immediate next step rather than a distant prospect. No controlled evaluation of task performance, safety or operational impact is reported in this paper.

| **Component** | **Status** | **Comment** |
|---|---|---|
| Tier 1 orchestrator: natural-language entry point, task decomposition, typed delegation, synthesis with cited execution trace | Implemented | Exercised in the site testing environment |
| Auditable decision-trace layer (Theodorakopoulos and Theodoropoulou, 2026; Radanliev, 2026) | Implemented | Running on FMS production history, rainfall feeds and workstation SOPs |
| Configurable autonomy levels: suggestion, approval and automatic | Implemented | Settable per scenario class and per agent |
| HSE rule checks within the validation stack | Reference design | Specified in Section 4.4, not yet enforced at run time |
| Long-term memory store and shift-handover retrieval | Reference design | Short-term context window only in the current prototype |
| Additional Tier 2 agents: maintenance, energy, safety and logistics | Future work | Enabled by the modular design of Section 4.9 |
| Multi-site federation through federated learning | Future work | Long-horizon direction |

Table 2a: Implementation Status of EI2ROC Architecture Components

| **Tool / Framework** | **Status in testing environment** | **Comment** |
|---|---|---|
| MMAU (Yin et al, 2025) | Exercised | Accounts for the Haiku-class calls of Table 3 |
| CORE (Michelakis, Hadjiyiannis and Stamoulis, 2025) | Implemented, not exercised | Hosted in a separate environment; not reachable from the orchestrator in the |
| | | current testing configuration |
| GEMMAS (Lee et al, 2025) | Exercised | Used for coordination diagnostics during testing |
| OpenAgentSafety (Vijayvargiya et al, 2025); hazard analysis (Dokas, 2026) | Active and blocking | Safety validation layer in the testing environment |
| Finance Agent Benchmark (Bigeard et al, 2025) | Exercised | Source of the cost-tracking methodology in Table 3 |
| Benchmark scenario library: the five scenario families of Section 5.4 | In progress | Under construction with site operations experts |
| Benchmarking harness: task generator, simulated IROC environment, tool emulators, trace logger, judge stack, metrics engine | Future work | Object of the calibration campaign of Section 7 |

Table 2b: Evaluation and Benchmarking Tools Mobilised in the Testing Environment
Table 2b lists the evaluation frameworks and benchmarking tools deployed alongside the prototype for validation purposes; they are not architectural components of EI2ROC.

### 4.11 Preliminary observations

The observations reported here characterise the operating envelope of the prototype. They do not constitute an evaluation: no baseline, no gold set and no human adjudication is involved, the sample comprises four dispatcher queries issued by a small number of users under a single model configuration, and no claim of statistical representativeness is made. Controlled measurement is the object of the calibration campaign of Section 7.

Across twelve weeks of testing, from 7 April to 30 June 2026 and 31 days of activity, total inference cost for the agent society was USD 19.63, which places the number of queries served over the period in the order of several dozen. Table 3 details four representative queries spanning a weekly performance review, a monthly review, a five-week trend analysis and a single-asset diagnostic. Costs are recomputed from logged token counts at published list prices.

Three observations follow. First, cost per query falls in a narrow band, from USD 0.28 to USD 0.42 with a median of USD 0.36, against production analyses whose downstream decisions Section 5.3 places in the millions of dollars. Second, end-to-end latency ranges from 70 to 130 s with a median of 90 s, consistent with the soft real-time, minutes-scale regime defined in Section 3.4, and it is dominated by the orchestrator's final synthesis calls rather than by tool execution, which returns in seconds. Third, the orchestrator's opening call carries an almost invariant 16 200 input tokens across queries, the fixed cost of its system prompt and tool schemas; prompt caching is not enabled in the current prototype, so this 16 200-token prefix is re-sent and re-billed on every query. At published Sonnet-class list prices (USD 3 per million input tokens), the prefix alone costs approximately USD 0.049 per query; enabling prompt caching, which applies a 90 per cent discount to cached tokens, would reduce this to approximately USD 0.005, saving roughly USD 0.044 per query. Extrapolated over the 40 to 70 queries implied by the total cost and the per-query band of Table 3, this represents a recoverable amount of USD 1.8 to 3.1, or approximately 10 to 15 per cent of the total USD 19.63 inference cost. Prompt caching is therefore an identified, low-effort optimisation rather than an inherent cost of the architecture, and its activation is scheduled before the calibration campaign of Section 7.

| Dispatcher query | Model calls (Sonnet / Haiku) | Input tokens | Output tokens | Latency (s) | Cost (USD) |
|---|---|---|---|---|---|
| Weekly performance review | 10 (4 / 6) | 110 050 | 7 727 | 70 | 0.33 |
| Monthly performance review | 8 (3 / 5) | 225 741 | 8 891 | 100 | 0.42 |
| Five-week trend analysis | 9 (3 / 6) | 144 585 | 13 025 | 130 | 0.39 |
| Single-asset diagnostic | 6 (3 / 3) | 87 516 | 7 500 | 80 | 0.28 |
| **Median** | **8.5** | **127 318** | **8 309** | **90** | **0.36** |

Table 3: Preliminary Observations on Four Dispatcher Queries in the Site Testing Environment

# BENCHMARKING INDUSTRIAL AGENTIC AI: STATE OF THE ART AND A PROPOSED FRAMEWORK

The architecture of Section 4 answers Gap (i) of Section 1.4, but an industrial agentic system reaches operational maturity only once it can be measured against the constraints that govern deployment. This section addresses Gap (ii) in three moves: what current agent evaluation offers, what it still misses for safety-critical industry, and the framework the authors propose to close that gap.

## 5.1 Why classical evaluation fails for agents

Outcome metrics (accuracy, precision, recall, F1) assume a static input–output mapping, whereas agents act through extended trajectories in which intermediate reasoning and action safety matter as much as the endpoint (Cao and Yu, 2026). An agent that reaches task success through unsafe tool calls or wasteful resource consumption is a deployment risk that outcome-only metrics cannot detect (Kapoor et al, 2024), and audits show that popular benchmarks systematically overestimate capability once holdout sets and realistic interaction patterns are enforced (Kapoor et al, 2024).

## 5.2 What exists: six dimensions and representative frameworks

Synthesising the evaluation literature, the authors assess an industrial agent along six orthogonal dimensions: outcome, path or trajectory, coordination, safety and constraint adherence, economic efficiency, and auditability or explainability (Table 4). A first generation of frameworks operationalises each. On outcome, MMAU evaluates five capability dimensions rather than task completion alone (Yin et al, 2025) and AgentBench attributes failures to long-term reasoning, decision-making and instruction-following weaknesses (Liu et al, 2024). On path, CORE encodes valid tool-use paths as deterministic finite automata and reports path correctness, prefix criticality and a harmful-call rate (Michelakis, Hadjiyiannis and Stamoulis, 2025), while AgentDiagnose scores backtracking, decomposition and self-verification against human judgment (Ou et al, 2025). On coordination, GEMMAS models agent interaction as a directed acyclic graph and derives an information-diversity score and an unnecessary-path ratio (Lee et al, 2025), and MultiAgentBench compares star, chain, tree and graph topologies under milestone-based scoring (Zhu et al, 2025); large-scale settings are covered by CREW-Wildfire under partial observability (Hyun, Waytowich and Chen, 2025) and by BenchMARL for reinforcement-learning baselines (Bettini, Prorok and Moens, 2024). On safety, OpenAgentSafety probes eight critical risk categories against real tools and reports unsafe behaviour in 51 to 73 per cent of safety-vulnerable tasks depending on the model (Vijayvargiya et al, 2025). On economics, the Finance Agent Benchmark tracks cost per query and finds the best model reaching 46.8 per cent accuracy at an average of USD 3.79 per query, with sharply diminishing returns beyond USD 1 (Bigeard et al, 2025). On auditability, recent work formalises evidence, decision-trace and outcome planes for operational decision-making (Theodorakopoulos and Theodoropoulou, 2026) and argues for traceability by design in agentic systems (Radanliev, 2026). Cross-cutting harnesses parallelise evaluation and curb overestimation (Kapoor et al, 2025; Kapoor et al, 2024). The community is thus converging on multi-dimensional evaluation, but each framework covers only a subset of dimensions, and a mining-side counterpart is absent: mining AI is still evaluated as isolated analytics using domain accuracy rather than as agentic systems (Guerin et al, 2025).

| Dimension | What it captures | Example metrics | Representative source(s) |
|---|---|---|---|
| Outcome | Success of the final task and quality of the produced answer | Task success rate, semantic correctness, end-state utility | MMAU (Yin et al, 2025) |
| Path | Validity of intermediate actions along the execution trajectory | Path correctness, harmful / useless tool calls, DFA-based path encoding | CORE (Michelakis, Hadjiyiannis and Stamoulis, 2025) |
| Coordination | Efficiency of inter-agent communication and division of work | Information-diversity score, ratio of redundant or useless paths, communication-graph diagnostics | GEMMAS (Lee et al, 2025) |
| Safety | Frequency and severity of unsafe or non-compliant actions | Harmful tool-call rate, goal-drift indicators, violation of operating envelopes | CORE (Michelakis, Hadjiyiannis and Stamoulis, 2025); AMDM (Shukla, 2026) |
| Economics | Computational and economic resources consumed per decision | Cost per query / per decision, end-to-end latency, throughput | Finance Agent Benchmark (Bigeard et al, 2025) |
| Auditability | Completeness of execution traces and compliance with governance requirements | Trace completeness, validator precision/recall, regulatory-coverage rate | POLARIS (Moslemi et al, 2026); AgentCompass (Kartik et al, 2025) |

Table 4: Six Evaluation Dimensions for Industrial Agentic AI Systems

### 5.3 What is missing: the industrial blind spot

The under-measurement of deployment-critical properties is now documented rather than merely asserted: a systematic review of agentic-AI evaluation finds that, of fifteen benchmarks examined, thirteen rely exclusively on binary task success, none score safety or security, and none include cost-efficiency in their primary evaluation (Kehkashan et al, 2026), while a parallel review reports that only 32 per cent of surveyed agentic systems document any safety mechanism (Yashwanth, Rishwanth and Shiny, 2026). Four constraints that govern CAPEX-intensive deployment remain under-addressed. First, latency reported against declared bounds: while some harnesses track execution duration, none reports it against a stated budget nor characterises its distribution, which is what sizing an industrial control loop requires (Kapoor et al, 2025; Farooq et al, 2026). Second, HSE compliance and fail-safe behaviour: general risk taxonomies (Vijayvargiya et al, 2025) do not engage lockout/tagout, permit-to-work or emergency-shutdown logic, and the industrial work that does engage hazard reasoning reports hallucination rates of 8 to 23 per cent in process-safety review (Dokas, 2026). Third, regulatory auditability: explainable traces (Radanliev, 2026;

Theodorakopoulos and Theodoropoulou, 2026) are not verified against IEC 61508, ISO 31000, IEC 62443 or ISA-95 (IEC, 2010; ISO, 2018; IEC, 2018; ISA, 2013). Fourth, cost-per-decision in capital-intensive operations, where a single dispatching choice may carry implications of several million dollars and must account for compute, API, human-oversight and opportunity costs. Safety-critical analogues in offshore oil and gas confirm that these four dimensions cannot be treated as optional add-ons (Shekhawat, Barua and Bhatia, 2025). Several pitfalls compound the gap: outcome myopia hides intermediate safety events; benchmark overfitting inflates apparent capability once holdout sets are enforced (Kapoor et al, 2024); accuracy reported without cost tracking obscures deployment infeasibility; and uncalibrated LLM-as-judge schemes drift, especially when judge and judged share architectural lineage (Zhuge et al, 2024; Kapoor et al, 2024). Together these define the evaluation gap that Gap (ii) formalises, and that the framework below is designed to close.

### 5.4 A proposed benchmarking framework

The authors propose a multi-dimensional benchmark for industrial agentic AI built on five principles: multi-dimensional (every run reports the six dimensions), process-aware (paths and coordination measured alongside outcomes), domain-grounded (scenarios derived from real open-pit telemetry rather than synthetic puzzles), human-calibrated (operations experts adjudicate a gold set), and reproducible/auditable (every run yields a complete, machine-readable trace) (Michelakis, Hadjiyiannis and Stamoulis, 2025; Farooq et al, 2026). The harness has six components (Figure 6) : a Task Generator (parameterised scenarios with controlled perturbations); a Simulated IROC Environment (emulated FMS, SCADA and IoT feeds, or a digital twin where available); Tool Emulators (deterministic and adversarial stubs); a Trace Logger (agent, tool, payload, latency, cost and safety flags, in an OpenTelemetry-compatible schema); a three-layer Judge Stack (deterministic validators, an Agent-as-a-Judge layer (Zhuge et al, 2024), and human adjudication for ambiguous cases); and a Metrics Engine that aggregates the six dimensions into an auditable scorecard. Each dimension maps to a concrete industrial metric, and the acceptance threshold attached to that metric is derived from the autonomy level at which the agent operates rather than fixed globally, because the cost of an error scales with the reversibility of the action it produces. Two thresholds are deductive and therefore invariant across autonomy levels: the harmful tool-call rate must be zero in HSE-critical scenarios, which follows from the fail-safe principle of IEC 61508 (IEC, 2010), and decision-trace logging must be complete, which follows from the reconstructibility requirement of ISO 31000 (ISO, 2018). The remaining thresholds are graduated. In suggestion mode, where a dispatcher adjudicates every output, criticality-weighted task success and DFA path correctness need only exceed the unaided-dispatcher baseline. In approval mode, where the human validates a recommendation without reconstructing the reasoning that produced it, both must be high enough for validation to remain meaningful rather than perfunctory, and the authors adopt 90 per cent and 85 per cent respectively as working values. These figures are anchored in two considerations. First, they are consistent with the SIL 2 integrity level of IEC 61508 (IEC, 2010), which tolerates a dangerous-failure probability of $10^{-3}$ to $10^{-2}$ per demand, implying that the vast majority of decisions must be correct for validation to remain meaningful. Second, they reflect the cognitive ergonomics of dispatcher validation: below approximately 85 per cent path correctness, a human reviewer must reconstruct the agent's reasoning from scratch to confirm safety, negating the time-saving purpose of the approval mode (Guerin et al, 2025). In automatic mode, restricted to bounded and reversible actions, the thresholds tighten towards the deductive ones. These working values are starting points to be fixed by the calibration panel described below, not normative constants, while cost per decision and p95 latency are budgeted per deployment against the value of the decisions concerned. The initial scenario library spans five families grounded in open-pit operations: weather-event response, critical equipment failure, Short-Interval-Control drift, capacity re-planning under uncertainty (Guerin et al, 2025), and adversarial/stress tests (failing tools, corrupted sensor feeds, prompt injection). Calibration follows a four-step, iterative protocol: a panel of dispatchers and superintendents annotates a gold set, inter-rater agreement is measured (Cohen's κ), and the automatic judges are calibrated against the gold set and re-run as new scenarios are added. Although grounded in mining, the design transfers to other CAPEX-intensive, HSE-regulated industries, namely offshore oil and gas (Shekhawat, Barua and Bhatia, 2025), energy dispatch and maintenance, manufacturing execution, and metallurgical process control, which share the same structure of heterogeneous subsystems, real-time constraints and regulated decisions. The framework is therefore positioned as a blueprint for industrial agentic-AI benchmarking, not a mining-only artefact.

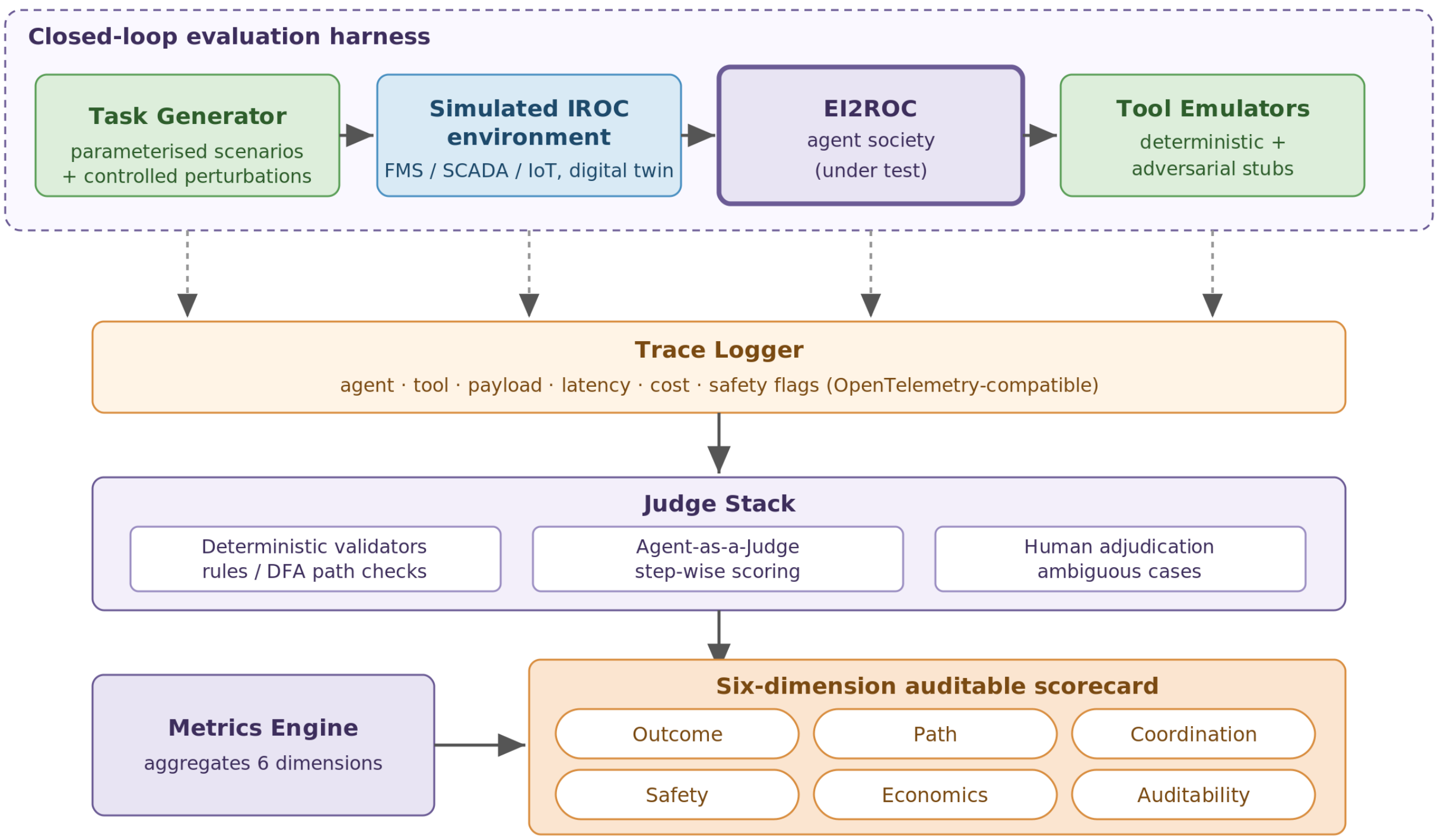

FIG 6: Reference architecture of the proposed benchmarking harness.

# DISCUSSION

## 6.1 Contributions revisited

The paper has presented four contributions: a synthesis of industry needs and documented gains in analogous industries (Section 3); the EI2ROC three-tier architecture, with its specialised agents, tool-call protocol, LLM-native routing and end-to-end query lifecycle (Section 4); a critical review of agentic-AI benchmarking along six dimensions; and a multi-dimensional benchmarking framework with scenario library and calibration protocol (Section 5).

## 6.2 Implications for industrial adoption

Three implications follow. Benchmarking is a deployment prerequisite, not a research nicety: industrial agentic systems cannot be promoted past the proof-of-concept stage without measurable cost-per-decision, HSE-compliance and regulatory-auditability evidence (IEC, 2010; ISO, 2018; IEC, 2018; ISA, 2013). The modular EI2ROC architecture lowers integration risk by enabling progressive deployment of specialised agents. And systematic traceability of agent decisions feeds the HSE-compliance and governance pipelines that mining operators are required to maintain.

### 6.3 Limitations

Two limitations bound the present contribution. First, the observations of Section 4.11 describe the operating envelope of the prototype on a handful of queries and are not a controlled evaluation; producing that evidence requires precisely the benchmarking framework of Section 5, whose calibration constitutes the next phase of this research. Second, the agentic-AI landscape evolves rapidly, so any benchmarking artefact must be designed for iterative re-calibration rather than as a static reference.

### 6.4 Threats to validity

Four threats to validity are explicit. Selection bias in the scenario library can shift the apparent performance of agents. LLM-judge fragility, documented in (Kapoor et al, 2024), must be controlled by inter-rater agreement and human re-anchoring. External validity beyond open-pit mining (eg, to underground operations) is asserted but not yet demonstrated. Finally, dependence on proprietary LLMs introduces cost, availability and vendor-lock-in risks that the framework's design (a modular tool registry and a model-agnostic protocol) mitigates but does not eliminate.

## FUTURE WORK

The immediate next step is a calibration campaign of the benchmarking framework against real scenarios, conducted with operations experts at the site, followed by an empirical benchmark of the agentic stack against human dispatchers on the scenario library instantiated on real telemetry. The additional agents and the multi-site federation listed as future work in Table 2a will be developed incrementally and evaluated against the same harness. Finally, a partial open-sourcing of the scenario library is planned, to support the emergence of a shared benchmarking practice across the industrial-AI community.

## CONCLUSION

This paper has addressed two intertwined gaps that constrain the evolution of mining IROCs towards full autonomy. The first gap, the absence of an integrated agentic architecture for mining IROCs, is answered by EI2ROC, a three-tier multi-agent design coupling an LLM orchestrator with specialised monitoring and production-capacity-prediction agents over a Model Context Protocol–style tool interface, with LLM-native routing, an auditable tool-call protocol and a dedicated data layer. The second gap, the absence of a benchmarking framework suited to safety-critical industrial agents, is answered by a multi-dimensional framework whose six evaluation dimensions, scenario library and human-calibration protocol are explicitly designed to bridge the industrial blind spot of current agentic benchmarks. Together, EI2ROC and the benchmarking framework constitute a methodological blueprint that the authors believe is transferable across CAPEX-intensive industries (offshore oil and gas, energy, manufacturing and metallurgy) where auditable agentic AI is poised to become a defining enabler of Industry 4.0.